\documentclass[10pt,twocolumn,letterpaper]{article}

\usepackage{cvpr}              

\usepackage[dvipsnames]{xcolor}

\usepackage[utf8]{inputenc} 
\usepackage[T1]{fontenc}    
\usepackage{url}            
\usepackage{booktabs}       
\usepackage{amsfonts}       
\usepackage{nicefrac}       
\usepackage{microtype}      
\usepackage{xcolor}         
\usepackage{scalerel}

\usepackage{times}
\usepackage{epsfig}
\usepackage{graphicx}
\usepackage{amsmath}
\usepackage{amssymb}
\usepackage{array,multirow}

\usepackage{graphicx}
\usepackage{amsmath}
\usepackage{amssymb}
\usepackage{booktabs}

\usepackage{rotating}

\usepackage{tabularx}
\usepackage{colortbl}
\usepackage{bm}
\usepackage{xspace}

\usepackage{enumitem}
\usepackage[olditem,oldenum]{paralist}

\usepackage[belowskip=1pt,aboveskip=5pt,font=small]{caption}
\usepackage[belowskip=0pt,aboveskip=3pt,font=small]{subcaption}
\newcommand{\xhdr}[1]{\vspace{3pt}\noindent\textbf{#1}\xspace}

\newcolumntype{s}{>{\columncolor[gray]{.85}[.5\tabcolsep]}c}

\makeatletter
\DeclareRobustCommand\onedot{\futurelet\@let@token\@onedot}
\def\@onedot{\ifx\@let@token.\else.\null\fi\xspace}

\usepackage[accsupp]{axessibility} 

\definecolor{cvprblue}{rgb}{0.21,0.49,0.74}
\usepackage[pagebackref,breaklinks,colorlinks,citecolor=cvprblue]{hyperref}

\usepackage[capitalize]{cleveref}
\crefname{section}{Sec.}{Secs.}
\Crefname{section}{Section}{Sections}
\Crefname{table}{Table}{Tables}
\crefname{table}{Tab.}{Tabs.}

\def\paperID{13137} 
\def\confName{CVPR}
\def\confYear{2024}

\title{SLAIM: Robust Dense Neural SLAM for Online Tracking and Mapping}

\author{
    Vincent Cartillier$^1$, Grant Schindler$^{2}$, Irfan Essa$^{1,2}$ \\
    $^1$Georgia Tech \quad $^2$ Google Research \\
    {\tt\small vcartillier3@gatech.edu} \\
    \tt\normalsize
    \href{https://vincentcartillier.github.io/slaim.html}{vincentcartillier.github.io/slaim.html}
}

\begin{document}

\twocolumn[{
\maketitle
\begin{center}
    \captionsetup{type=figure}
    \includegraphics[width=0.95\textwidth]{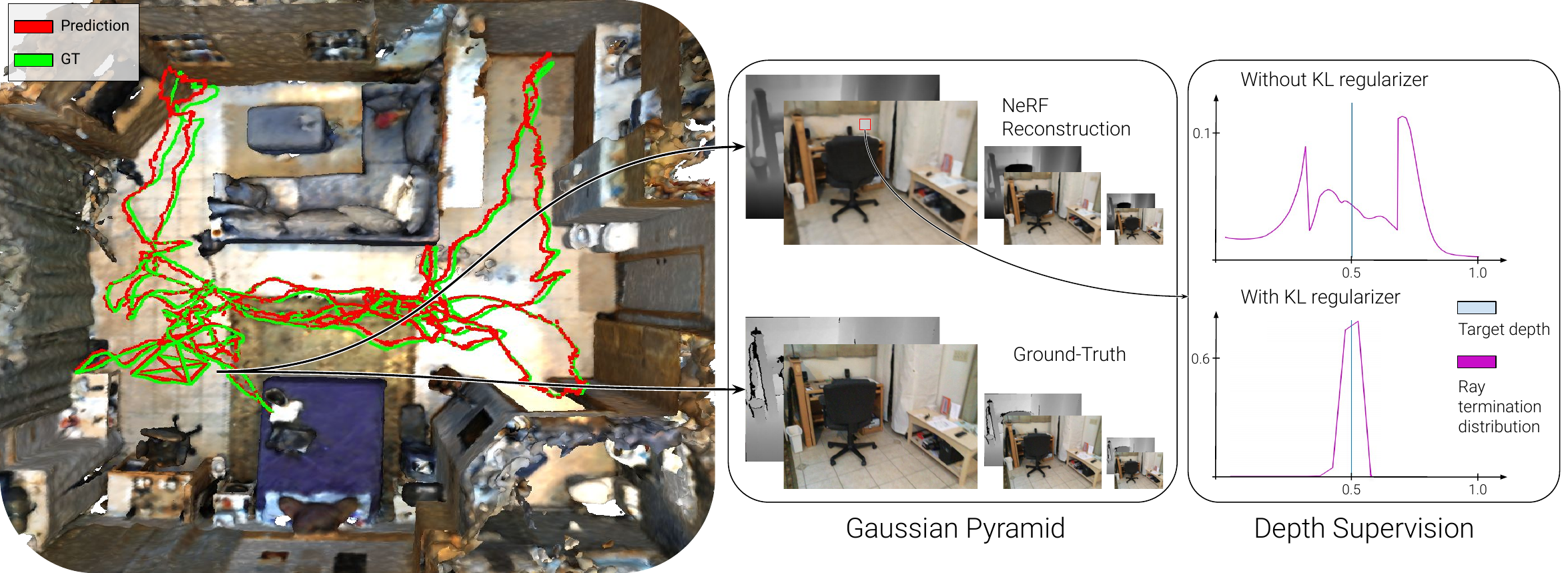}
    \captionof{figure}{ 
    We present SLAIM, a robust dense neural RGB-D SLAM system that performs online tracking and mapping in real time. 
    SLAIM implements a Gaussian Pyramid filter on top of NeRF to perform coarse-to-fine tracking and mapping. We also introduce a new target ray termination distribution that we use in a KL regularizer to supervise the network to converge towards the right geometry. SLAIM reaches state-of-the-art results in both tracking and 3D reconstruction accuracy. }
    \label{fig:teaser}
\end{center}
}]




\begin{abstract}

We present SLAIM - Simultaneous Localization and Implicit Mapping. We propose a novel coarse-to-fine tracking model tailored for Neural Radiance Field SLAM (NeRF-SLAM) to achieve state-of-the-art tracking performance. 
Notably, existing NeRF-SLAM systems consistently exhibit inferior tracking performance compared to traditional SLAM algorithms \cite{mur2015orb}.
NeRF-SLAM methods solve camera tracking via image alignment and photometric bundle-adjustment. Such optimization processes are difficult to optimize due to the narrow basin of attraction of the optimization loss in image space (local minima) and the lack of initial correspondences. We mitigate these limitations by implementing a Gaussian pyramid filter on top of NeRF, facilitating a coarse-to-fine tracking optimization strategy. 
Furthermore, NeRF systems encounter challenges in converging to the right geometry with limited input views. While prior approaches use a Signed-Distance Function (SDF)-based NeRF and directly supervise SDF values by approximating ground truth SDF through depth measurements, this often results in suboptimal geometry. In contrast, our method employs a volume density representation and introduces a novel KL regularizer on the ray termination distribution, constraining scene geometry to consist of empty space and opaque surfaces.
Our solution implements both local and global bundle-adjustment to produce a robust (coarse-to-fine) and accurate (KL regularizer) SLAM solution. 
We conduct experiments on multiple datasets (ScanNet, TUM, Replica) showing state-of-the-art results in tracking and in reconstruction accuracy.




\end{abstract}

\section{Introduction}
\label{sec:intro}


Dense visual Simultaneous Localization and Mapping (SLAM) is a long-standing problem in 3D computer vision with many applications, including autonomous driving, indoor and outdoor robotic navigation, virtual reality, and augmented reality. In this work, we present state-of-the-art tracking results using implicit maps to improve SLAM.

Traditional SLAM systems start by estimating image correspondences, which can be sparse, in the form of matched keypoints \cite{mur2017orb} between frames, or dense \cite{teed2021droid} via estimation of optical flow for instance. These correspondences are then further used in a bundle adjustment process to predict and refine camera poses. The ability to find such correspondences is a large prior condition and assumption for traditional SLAM systems to work. This assumption may fail under some circumstances. In the sparse case, it is not always easy to detect keypoints -- for instance texture-less surfaces are difficult to track. In the dense case, a deep pre-trained neural network is usually used, which limits the applications to scenes covering similar statistics as the ones used in the training set.

An emerging direction of research has presented new methods to solve SLAM using implicit maps (NeRFs), thus removing the requirement of correspondences in the pipeline. This new class of methods is best described as \emph{NeRF-SLAM}. These NeRF-SLAM methods are interesting because they define SLAM as a full optimization problem without the need of any external pre-computed information (i.e., keypoints or a pre-trained network). The general idea is to (1) map past frames by building a NeRF model and (2) use the view synthesis capabilities of the NeRF model to track new frames via image alignment and photometric bundle-adjustment. iMAP \cite{sucar2021imap} was the first implementation to demonstrate the feasibility of this approach. NICE-SLAM \cite{zhu2022nice} improved the results by using a more accurate NeRF model producing better synthetic views. 
Both iMAP and NICE-SLAM represent the geometry as a volume density in NeRF \cite{mildenhall2021nerf}.

More recently, ESLAM \cite{johari2022eslam} and Co-SLAM \cite{wang2023co} represent the geometry using a signed-distance function (SDF) and show better tracking and 3D reconstruction performances. Both solutions directly apply supervision to the SDF predicted values from NeRF which enforces the ray termination distribution to be unimodal and centered on the depth measurement. This leads to a more efficient and performant solution.
However, the SDF supervision is done by computing an approximation of the ground-truth SDF values using the depth measurement. This approximation leads to sub-optimal geometry convergence. 

In our proposed work, we keep the original volume density definition of NeRF but we apply a KL regularizer over the ray termination distribution to constraint them to be a narrow unimodal distribution. This effectively improves efficiencies and performances while not restricting the underlying geometry to a sub-optimal solution.
In addition, ESLAM and Co-SLAM solve SLAM with local and global bundle-adjustment respectively. Instead we present a solution that combines the best of both worlds by performing local and global bundle adjustment. 

A common observation on all previous NeRF-SLAM baselines is the gap in terms of tracking performances when compared to more traditional SLAM systems \cite{mur2017orb}.
A key problem with these baselines is that any high spatial frequencies contained in accurate NeRF renderings actually make the image alignment optimization step difficult and inefficient to solve.  The high frequencies induce a narrower basin of convergence during the image alignment process. 

In this paper we present a new coarse-to-fine NeRF-SLAM pipeline to overcome this image alignment problem, achieving state-of-the-art results. The core idea is to use a Gaussian filter on the output image signal to enlarge the basin of attraction during image alignment optimization and photometric bundle-adjustment, making the tracking more robust and efficient.  This is a similar idea to the classical hierarchical Lucas-Kanade optical flow algorithm~\cite{lucas1981iterative} that solves optical flow using image pyramids~\cite{burt1987laplacian}.  In the NeRF-SLAM context, applying this central idea requires overcoming key technical hurdles that we present below.

Overall our contributions are as follows:
\begin{itemize}
    \item We present SLAIM, a robust, NeRF-SLAM system using implicit mapping and a coarse-to-fine improved tracking.
    

    \item We present a new KL regularizer over the ray termination distribution which leads to better tracking and mapping. 

    \item We present a new NeRF-SLAM pipeline that performs both local and global bundle-adjustment.
    
    \item We report state-of-the-art results on camera pose predictions and 3D reconstructions on several datasets.
\end{itemize}

\section{Related Work}
\label{sec:related_work}

\subsection{Visual SLAM}
The goal of visual SLAM is to estimate camera poses along with a global 3D reconstruction given a stream of input frames \cite{cadena2016past}. Most solutions to this problem implement a mapping and tracking module working in parallel \cite{klein2009parallel}. Traditionally, tracking is solved by iteratively finding correspondences between frames and updating the poses accordingly \cite{mur2015orb, mur2017orb, campos2021orb, rosinol2020kimera, teed2021droid}. These correspondences can be found by image keypoints matching \cite{mur2015orb, mur2017orb, campos2021orb, rosinol2020kimera}. More recently Droid-SLAM \cite{teed2021droid} estimates correspondences from dense deep correlation of features. All of the previous work are indirect SLAM methods and performances rely on good keypoints features \cite{mur2015orb, mur2017orb, campos2021orb, rosinol2020kimera} or pre-trained networks \cite{teed2021droid}. Our proposed pipeline, a direct SLAM solution, does not require any pretraining or pre-processing.

We focus on dense reconstruction mapping (i.e., dense-SLAM) where a dense 3D map is maintained during the entire tracking process \cite{newcombe2011kinectfusion, newcombe2011dtam}. Prior work used a fixed or a hierarchy of resolutions \cite{newcombe2011kinectfusion, dai2017bundlefusion,schops2019bad, tang2020deep}. These approaches are memory inefficient and are limited in the level of details represented. Other works \cite{dai2020sg,chabra2020deep,peng2020convolutional} use training methods to improve the 3D reconstruction. However, these methods require extra training data that is not available in our problem. Closer to our work we find approaches that store information in a world-centric map representation using surfels \cite{schops2019bad, whelan2015elasticfusion} or voxels \cite{newcombe2011kinectfusion,bylow2013real,dai2017bundlefusion,kahler2016real, zhu2022nice}. Our work stores implicit latent features in a hashmap of voxels.

\subsection{Neural Radiance Fields (NeRFs)} 
Scene representations and graphics have recently made substantial progress by using a neural network MLP combined with volumetric rendering to allow for novel view synthesis \cite{mildenhall2021nerf}. NeRF \cite{mildenhall2021nerf} has triggered a large number of subsequent papers that proposed improvements to the initial model. 
\cite{zhu2022nice,takikawa2021neural,fridovich2022plenoxels,muller2022instant} drastically improved the training time by using a smaller MLP combined with a set of optimizable spatial grid features. Inspired by this performance, we build our work upon Instant-NGP \cite{muller2022instant} which uses a hash-map of hierarchical grid representations of the scene allowing real-time training.
In another line of work, we find papers that alleviate the dependency of NeRF on having known camera poses. \cite{wang2021nerf} adds camera poses as parameters in the overall optimization pipeline. The overall process is very slow. 

BARF \cite{lin2021barf} shares a similar idea to our work as it solves bundle adjustment using a coarse-to-fine image alignment optimization setup. The system modulates the NeRF's positional encoding using a low-pass filter which effectively reduces high frequencies and widens the basin of convergence. Since we are using grid features as positional encoding we cannot apply the same technique. Moreover,
BARF assumes all frames with initial noisy poses given at the beginning of the optimization which is not compatible with an on-line SLAM formulation. 
In terms of scene representation, the initial NeRF \cite{mildenhall2021nerf} characterizes scene geometry through volume density. Alternatively, some researchers have explored the use of a signed-distance function (SDF) to encode geometry \cite{yariv2021volume,yariv2023bakedsdf,or2022stylesdf,azinovic2022neural,wang2021neus}. While this approach can allow direct supervision over SDF values \cite{johari2022eslam,wang2023co}, it necessitates intricate techniques for unbiased conversion of SDF to ray termination probability \cite{wang2021neus}.
In our investigation, we encounter DS-NeRF \cite{deng2022depth}, which uses the original density definition and imposes a Gaussian KL regularization on the ray termination distribution to ensure unimodality. 
In our work we apply regularization differently by modeling a skewed Gaussian-like distribution which is shown to be more accurate.

\subsection{SLAM with NeRF}
In the NeRF-SLAM domain we start by finding \cite{rosinol2022nerf} that combines Droid-SLAM \cite{teed2021droid} and Instant-NGP \cite{muller2022instant}. However, \cite{rosinol2022nerf} solves SLAM without NeRF. 
Closer to our work, we find SLAM implementations using implicit mapping, including ESLAM, Co-SLAM and others \cite{zhu2022nice, sucar2021imap, kong2023vmap, wang2023co, johari2022eslam}. 
ESLAM and Co-SLAM are using an SDF-based NeRF and add supervision over the SDF values directly. 
Supervision is done via computing an approximate ground-truth SDF value given a ray depth measurement. This approach provides improved control over scene geometry and accelerates convergence. Nevertheless, this approximation may result in suboptimal geometry. In contrast, our proposed approach incorporates a KL regularizer on the ray termination distribution, achieving an optimal and rapid convergence.
None of the previous techniques study coarse-to-fine approaches. 


\section{Methodology}
\label{sec:method}







\begin{figure*}[h]
    \centering
    \includegraphics[width=0.95\textwidth]{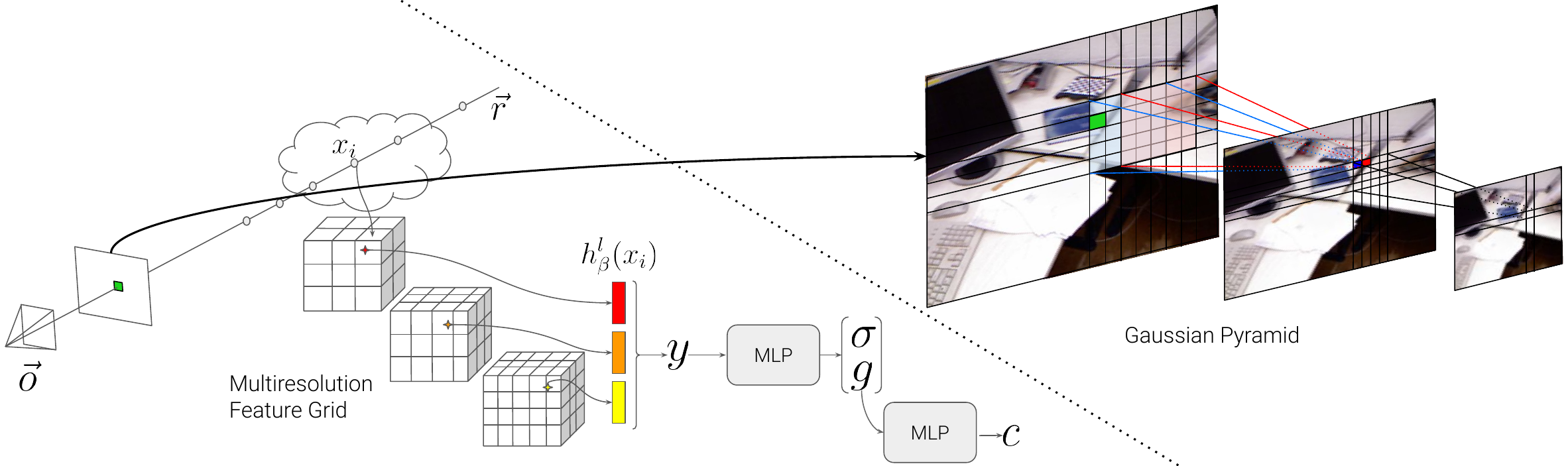}
    \caption{\textbf{Overview of SLAIM.} For a given input ray with center $\vec{o}$ and direction $\vec{r}$ we start by rendering its corresponding color pixel via ray tracing and volume rendering. For each sample $x_i$ along the ray we query the multiresolution feature grid to form an input embedding $y$. We then make successive calls to two shallow MLP networks to predict a density $\sigma$ and color $c$ for that sample. After reconstructing the image we apply a Gaussian Pyramid filter to perform coarse-to-fine tracking and mapping.}
    
    \label{fig:model}
\end{figure*}

We provide an overview of our method in Fig. \ref{fig:model}. 
In this section we detail SLAIM, a novel approach for dense mapping and tracking of an RGB-D input $\{I_t\}_{t=1}^N$ stream using known camera intrinsics $K \in \mathbf{R}^{3\times 3}$ and a neural scene representation $f_{\Psi}$. The camera poses $\{P_t\}_{t=1}^N$ and implicit scene representation are jointly optimized in a coarse-to-fine manner.

\subsection{NeRF pre-requisites}

 The geometry is encoded in a multiresolution hash-based feature grid \cite{muller2022instant}, with parameters $\beta$, that maps a 3D input coordinate $x \in \mathbf{R}^3$ into a dense feature vector $y = h_{\beta}^{L}(x)$. The feature space is divided into $L$ grids of resolution ranging from $R_{min}$ to $R_{max}$. Feature vectors at each level $l \in [1,L]$,  $h^{l}_{\beta}(x)$ are queried via trilinear interpolation and further concatenated to form the final positional encoding $y = \left[h^{1}_{\beta}(x), ..., h^{L}_{\beta}(x) \right]$. We further employ two shallow MLP decoders to estimate the density and color at the given 3D input location. Specifically, the geometry decoder, with parameters $\tau$, outputs a density value $\sigma$ and a feature vector $g$. 
The color decoder, with parameters $\psi$, takes as input the feature vector $g$ and predicts RGB values $c$.
\begin{equation}
f^{g}_{\tau}(y) \mapsto (\sigma, g)
\text{;}\quad
    f^{c}_{\psi}(g) \mapsto c
\end{equation}

Note that in traditional NeRF systems \cite{mildenhall2021nerf} the color decoder also takes as input the direction embedding. Here we intentionally remove that input to reduce the optimization complexity. This choice is driven by our focus on camera tracking and mapping, with no intention of modeling specularities. 
During mapping we optimize the set of NeRF parameters $\Psi=\{\beta, \tau, \psi\}$ along with the camera poses parameters.
Following \cite{mildenhall2021nerf, muller2022instant} we render color and depth pixels by alpha compositing the values along a ray. 
Specifically, given the camera origin $\vec{o} \in \mathbf{R}^3$ and ray direction $\vec{r} \in \mathbf{R}^3$, we sample $M$ points $x_i = \vec{o} + d_i.\vec{r}$, with $i \in \{1,...,M\}$ and depths values $d_i$.
We employ the importance sampling strategy implemented in \cite{muller2022instant} to sample points along a ray. We bound the scene to the unit cube $[0,1]^3$ and we start by uniformly sampling with a fixed ray marching step size of $\Delta_r=\sqrt{3}/1024$.
In addition, we maintain an occupancy grid to avoid sampling points in free spaces, and we stop sampling once a surface is found.
Similar to \cite{mildenhall2021nerf}, we model the ray termination distribution $w(r)$ as follow:
\begin{equation} \label{eq:ray_dist}
    w(r) = T(r) \sigma(r)
\end{equation}
 with $T(r) = exp(-\int_{0}^{r} \sigma(s) ds)$. We approximate this integral using a sampling-based method and express the discretized $w_i$ at point $x_i$ as $w_i=\alpha_i.T_i$ with $\alpha_i = (1 - e^{-\sigma_i \delta_t})$ and $T_i=\prod_{j=1}^{i-1} (1-\alpha_i)$. The final pixel color and depth values are computed as follows:

\begin{equation} 
    \hat{c} = \sum_{i=1}^{M} w_i.c_i 
    \text{;}\quad
    \hat{d} = \sum_{i=1}^{M} w_i.d_i 
\end{equation}

\subsection{Depth Supervision}
\begin{figure*}[h]
    \centering
    \vstretch{.9}{
    \includegraphics[width=0.95\textwidth]{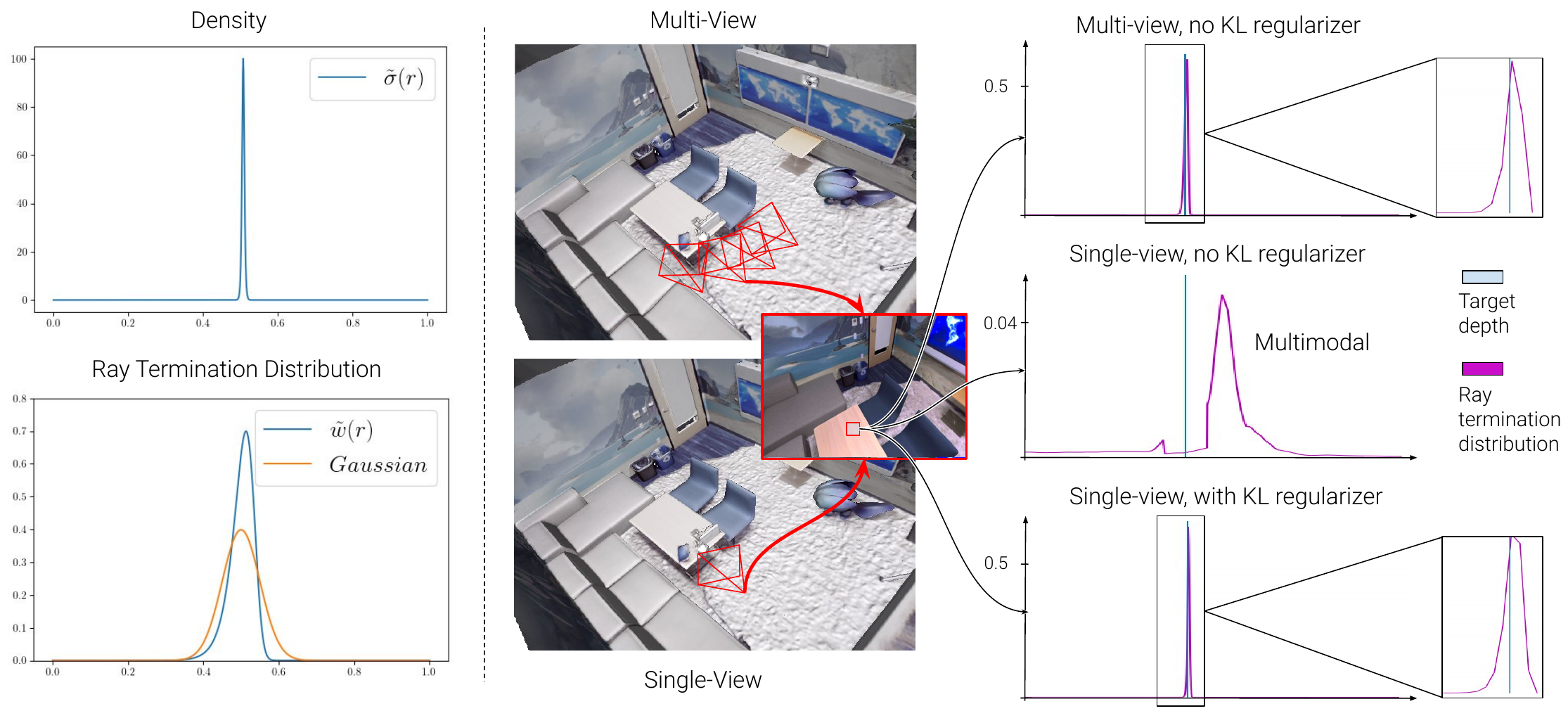}
    }
    \caption{Illustration of the custom ray termination distribution. (left-top) shows the estimation of the density response $\tilde{\sigma} \sim sech^2$ as a narrow bell-shape function. (left-bottom) compares the resulting ray termination distribution $\tilde{w}$ computed from $\tilde{\sigma}$ to the Gaussian distribution described in DS-NeRF \cite{deng2022depth}. 
    On the right, we display results from an experiment showcasing the ray termination distribution' shape under different conditions. (middle) compares mapping setups: one with multiple training views (over-constrained geometry) and another with only one view (under-constrained geometry). (right-top) shows multi-view training without KL regularization (right-middle) depicts single-view training without KL regularization, and (right-bottom) exhibits single-view training with our custom regularizer. We observe similarity between the ray termination distributions in the (right-top) and (right-bottom), supporting the use of our custom distribution.}
    \label{fig:kl_reg}
\end{figure*}

Using depth supervision is crucial for a NeRF-SLAM system to produce the right geometry. Using a loss (e.g L1) over the depth values directly can lead to reconstruction artifacts in regions with only limited views. This is because the volumetric rendering does not adhere to the restriction that the majority of the geometry consists of empty space and opaque surfaces \cite{deng2022depth}. In other words, the ray termination distribution (i.e. the probability of a ray to hit an obstacle) should be unimodal and centered on the depth measurement. 
We implement a new KL regularizer over these ray termination distributions to enforce them to be unimodal. The ideal distribution is a delta function centered at the depth measurement $d$: $\delta(r-d)$. DSNeRF \cite{deng2022depth} approximates this distribution with a narrow Gaussian distribution. 
Nevertheless, we observe that the distribution arising from Eq. \ref{eq:ray_dist} closely resembles a skewed normal distribution, as illustrated in Fig. \ref{fig:kl_reg}. This resemblance is attributed to the fact that a depth measurement implies a spike, or a narrow bell-shaped, \textit{density} response $\sigma(r)$ centered at the depth. Consequently, Eq. \ref{eq:ray_dist} yields a skewed Gaussian-like ray termination distribution.
Therefore, we approximate an ideal density function $\tilde{\sigma}$ and compute the ray termination distribution $\tilde{w}$ using Eq. \ref{eq:ray_dist}. We define $\tilde{\sigma}(r) =S\times sech^2(\frac{(r-d')}{\sigma_d})$ with $d'$ and $\sigma_d$ the mean and variance and $S$ a scale factor. This function is a narrow bell-shape function centered on $d'$. We choose the $sech^2$ function instead of a Gaussian for its mathematical simplicity when computing integrals.
We set the $d'$ parameter such that the expected depth $\hat{d}$ matches the depth measurement $d$, $E_x(\tilde{w})=d$. 
The final KL regularizer function is expressed as follows:
\begin{equation}
\begin{split}
      \mathcal{L}_{KL} & =  -\frac{1}{N}\sum^{N}_{n=1} \sum_k log(w(r_k))\tilde{w}(r_k) \Delta_r
\end{split}
\end{equation}
We show in the supplement how to compute the expectation and how to derive such KL loss.

\subsection{Coarse-to-fine Tracking and Mapping.}

\xhdr{Coarse-to-fine formulation.}
Our system, similar to previous ones \cite{sucar2021imap, zhu2022nice, wang2023co}, tracks cameras via image alignment and photometric bundle adjustment. 
Achieving good camera pose estimations given this approach can be difficult due to the narrow basin of attraction \cite{lin2021barf, newcombe2011dtam} of the optimization function (i.e., there are many local minima in image space) and the lack of initial correspondences. To overcome this problem we use a coarse-to-fine approach using a Gaussian Pyramid filter \cite{burt1987laplacian} to effectively smooth the input signal in the early iterations in order to widen the basin of attraction and thus avoid getting stuck in local minima. 

To render an image at a level $l_G$ in the Gaussian Pyramid we perform multiple reduce operations \cite{burt1987laplacian} which consists of applying a convolution filter with a 5-tap kernel $w_{g}= [1,4,6,4,1] \times [1,4,6,4,1]^T / 256$ and downsampling the results by selecting every two pixels. We apply a similar coarse-to-fine approach on the depth frame, but instead of a 5-tap kernel we apply a median filter.
We apply this Gaussian pyramid filter on both reconstructed and ground-truth images. Reconstructing an entire image using NeRF during optimization would be impractical due to memory and time complexity constraints. Instead we sample a set of pixels at the Gaussian Pyramid level $l_G$ and compute the corresponding receptive field on the original image. We then restrict the NeRF pixel reconstruction to the pixels within the receptive field. 
We can derive the steepest descent image formulation of the smooth image $I^{(l_G)}$ after applying the Gaussian Pyramid filter with $l_G$ levels.

\begin{equation}
    J(I^{(l_G)}, P) = \prod_{k=1}^{l_G} \frac{\partial I^{k}}{\partial I^{k-1}}  \frac{\partial I^{0}}{\partial P}
\end{equation}

Where $\partial I^{0} / \partial P$ is the partial derivative of the full resolution image given the camera pose parameters $P$, and $\partial I^{k} / \partial I^{k-1}$ is the Jacobian matrix between images at levels $k$ and $k-1$ in the Gaussian Pyramid. These matrices are populated with the kernel weights $w_g$. From this formulation this is akin to applying a weighted average on image gradients $\partial I^{0} / \partial P$. This has the effect to smooth and cohere the gradients for a more robust optimization.


\xhdr{Camera Tracking.} Similar to previous works \cite{zhu2022nice, sucar2021imap, wang2023co} our tracking strategy is defined as an image alignment problem between the current frame and the underlying implicit map. Optimization is performed via minimizing a photometric and geometrical loss along with our KL regularizer.
\begin{equation}\label{eq:loss}
  \mathcal{L} = \mathcal{L}_{rgb} + \lambda_{d} \mathcal{L}_{d} + \lambda_{KL} \mathcal{L}_{KL}   
\end{equation}
The color loss $\mathcal{L}_{rgb}$ is an $l_2$ loss over the RGB pixel values and the depth loss $\mathcal{L}_{d}$ is an $l_1$ loss over depth values. 
\begin{equation}
    \mathcal{L}_{rgb} = \frac{1}{N}\sum_{n=1}^{N}(\hat{c}_n - c_n)^2
    \text{;            }
    \mathcal{L}_{d} = \frac{1}{|\mathcal{I}_d|}\sum_{r \in \mathcal{I}_d} |\hat{d}_r - d_r|
\end{equation}

Where $\mathcal{I}_d$ is the subset of sampled rays with a valid depth measurement. 
Using the defined loss we track the camera-to-world pose matrix $P_t \in SE(3)$ for each timestep $t$. We initialize the pose of the current frame using constant speed assumption $P_t = P_{t-1} P_{t-2}^{-1} P_{t-1}$.
The tracking optimization consists of doing multiple iterations of selecting $N=N_t$ pixels within the current frame and optimizing the pose by minimizing the tracking loss via stochastic gradient descent.

\xhdr{Mapping.} 
We continuously optimize the scene representation with a set of selected keyframes. Similarly to \cite{wang2023co} we select keyframes at fixed intervals.
Given the set of growing keyframes $\mathcal{K}$ we jointly optimize the scene parameters $\Psi$ and camera poses $P_k$, $k \in \mathcal{K}$ using the same loss $\mathcal{L}$ from Eq. \ref{eq:loss}. 
 The joint optimization occurs through alternating steps. In each iteration, we optimize the scene representation $\Psi$, however the gradients of camera poses are accumulated over $k_p$ iterations before updating these parameters.
Mapping is performed every X frames and that process is split into two phases.
We start by running local bundle adjustment over the $M_{\mathcal{K}}-1$ most recent keyframes and the current frame, and then continue by running additional global bundle adjustment iterations over all keyframes $\mathcal{K}$.
We observed that when only doing global bundle adjustment, the quality of image reconstruction decreases as the number of keyframe increases. This is because the number of sampled rays per frame then reduces. Adding, local bundle adjustment helps maintaining a good reconstruction quality throughout the video which is crucial for tracking performances.




\subsection{Implementation details}
The proposed approach follows a two-fold process of tracking and mapping. To initialize the system, a few mapping iterations are executed on the first frame. Subsequently, for each new frame, we run the tracking and mapping processes in a sequential manner. 
The entire algorithm is built upon Instant-NGP \cite{muller2022instant} and is written in C++ and CUDA kernels leading to fast computation. We conduct experiments on a single NVIDIA A40 GPU and 2.35GHz AMD EPYC 7452 32-Core CPU. For experiments with default settings (Ours) we sample $N_t=1024$ and $N_m=2048$ pixels for tracking and mapping over $12$ and $30$ iterations respectively. Refer to the supplementary material for additional details.





\section{Experiments}
\label{sec:experiments}
\subsection{Experimental setup.}

\xhdr{Datasets.} We evaluate SLAIM on different scenes from three different datasets. Similar to previous work, we compare reconstruction performance on 8 synthetic scenes of the Replica dataset \cite{straub2019replica}. We evaluate the tracking performance of SLAIM on 6 scenes of the ScanNet \cite{dai2017scannet} dataset and 3 scenes from TUM-RGBD \cite{sturm2012benchmark}.

\xhdr{Metrics.} Following previous baselines \cite{zhu2022nice,wang2023co,johari2022eslam}, we measure the reconstruction quality on observed regions (within camera FoV) using Accuracy (cm), Completion (cm), Depth L1 (cm), and Completion ratio ($\%$) with a threshold of 5cm. We evaluate camera tracking using absolute trajectory error (ATE) RMSE \cite{sturm2012benchmark} (cm).

\xhdr{Baselines.} We consider iMAP \cite{sucar2021imap}, NICE-SLAM \cite{zhu2022nice}, Co-SLAM \cite{wang2023co} and ESLAM \cite{johari2022eslam} as baselines for comparison of reconstruction quality and camera tracking. 
We also compare to another version of SLAIM, that we refer to as $SLAIM_{MG}$ ($MG$ stands for max-grid), which renders images using different grid-resolution features to perform coarse-to-fine rendering instead of using a Gaussian Pyramid. Specifically, we use a max-grid-level parameter $mgl \leq L$ that we use to set a max resolution the network is allowed to use to construct the position embedding $y$. Grid features for higher levels are set to $0$, and the final positional encoding can be written as $y = \left[h^{1}_{\beta}(x), ..., h^{mgl}_{\beta}(x), 0 \right]$. This is a very simple trick technique to effectively render blurry images. See supplementary material for visualizations.

\xhdr{Coarse-to-fine settings.} 
In all of our experiments we apply the coarse-to-fine strategy to both tracking and mapping. We set an initial Gaussian Pyramid level (GPL) and gradually reduce that GPL throughout a mapping or tracking phase.
For a given number of iterations and initial GPL, we split evenly the number of iterations per pyramid levels. 
For example, in Replica we run 20 mapping iterations with an initial GPL of 1. This means we run 10 iterations at levels 1 and 0 respectively (level 0 means no blurring). 


\subsection{Tracking and Reconstruction performance.}
\begin{figure}[!ht]
    \centering
    \includegraphics[width=0.95\columnwidth]{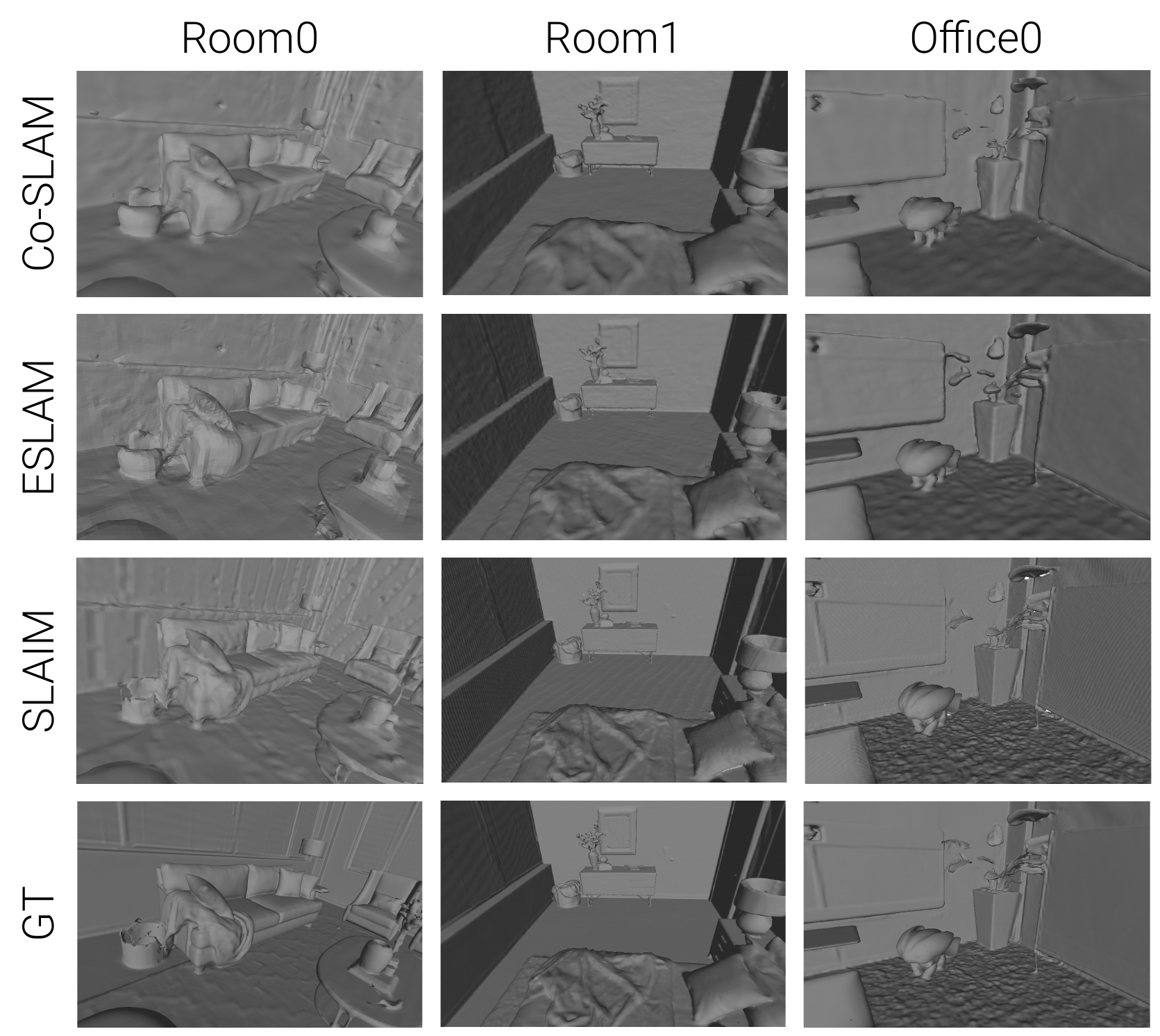}
    \caption{Reconstruction results on Replica \cite{straub2019replica}. Our method can retrieve thin structures where other baselines tend to oversmooth the geometry.
    }
    \label{fig:results_rec_figure}
\end{figure}

\textbf{Replica dataset.} We evaluate 3D reconstruction performance on the same simulated RGB-D sequences as iMAP. We report numbers in Tab. \ref{tab:results_replica} and qualitative examples in Fig. \ref{fig:results_rec_figure}.
Prior to running the evaluation we follow the implementation in Co-SLAM \cite{wang2023co} and perform mesh culling to remove unobserved regions outside of any camera frustum.
As shown in Tab. \ref{tab:results_replica}, our method achieves the best results in terms of accuracy with an improvement of close to $5\%$ compared to Co-SLAM \cite{wang2023co} and ESLAM \cite{johari2022eslam}. 
On the Depth-L1 and completion metrics, SLAIM performs second best. SLAIM struggles in estimating completely unobserved regions. This is not surprising given that NeRF cannot hallucinate large unobserved regions. Mesh holes in missing regions will lead to large Depth-L1 and completion scores. 
We additionally evaluate against two baselines: $SLAIM_{noKL}$, which omits the KL regularizer, and $SLAIM_{G}$, leveraging a Gaussian KL regularizer, as per DSNerf \cite{deng2022depth}. We observe that $SLAIM_{noKL}$ performs worse than NICE-SLAM \cite{zhu2022nice}. While both approaches employ a density-based NeRF with multi-resolution feature grids, NICE-SLAM utilizes pretrained decoders that incorporate learned geometrical priors. In contrast, our method is trained from scratch. Consequently, NICE-SLAM achieves superior reconstruction results. Comparing SLAIM to $SLAIM_{G}$, we consistently observe performance improvement by employing our custom ray termination distribution instead of a Gaussian. This suggests that $\tilde{w}(r)$ helps at better represent the geometry.

\begin{table}[ht]
\centering
\renewcommand{\arraystretch}{1.15}
\resizebox{0.95\columnwidth}{!}{
\begin{tabular}{ l c c c c c  }
\toprule
& & Depth L1(cm) $\downarrow$ & Acc. (cm) $\downarrow$ & Comp. (cm) $\downarrow$ & Comp. ratio (cm) $\uparrow$\\ 
\midrule
iMAP\cite{sucar2021imap} & & 4.64 & 3.62 & 4.93 & 80.51 \\
NICE-SLAM \cite{zhu2022nice} & & 1.90 & 2.37 & 2.64 & 91.13 \\
Co-SLAM \cite{wang2023co} & & 1.51 & 2.10 & 2.08 & 93.44  \\
ESLAM \cite{johari2022eslam} & & \textbf{0.94} & 2.18 & 
\textbf{1.75} & \textbf{96.46}  \\
\midrule
$SLAIM_{noKL}$ & & 2.13 & 2.45 & 2.78 & 89.17 \\
$SLAIM_{G}$ & & 1.46 & 2.11 & 1.98 & 94.87 \\
$SLAIM$ ($Ours$) & & 1.37 & \textbf{1.82} & 1.90 & 95.61 \\
\bottomrule
\end{tabular}}\\[2pt]
\caption{Reconstruction results on the Replica dataset \cite{straub2019replica}. SLAIM is best on accuracy and second best on all other metrics.
}
\label{tab:results_replica}
\end{table}

\begin{figure*}[!htb]
    \centering
    \includegraphics[width=1.\textwidth]{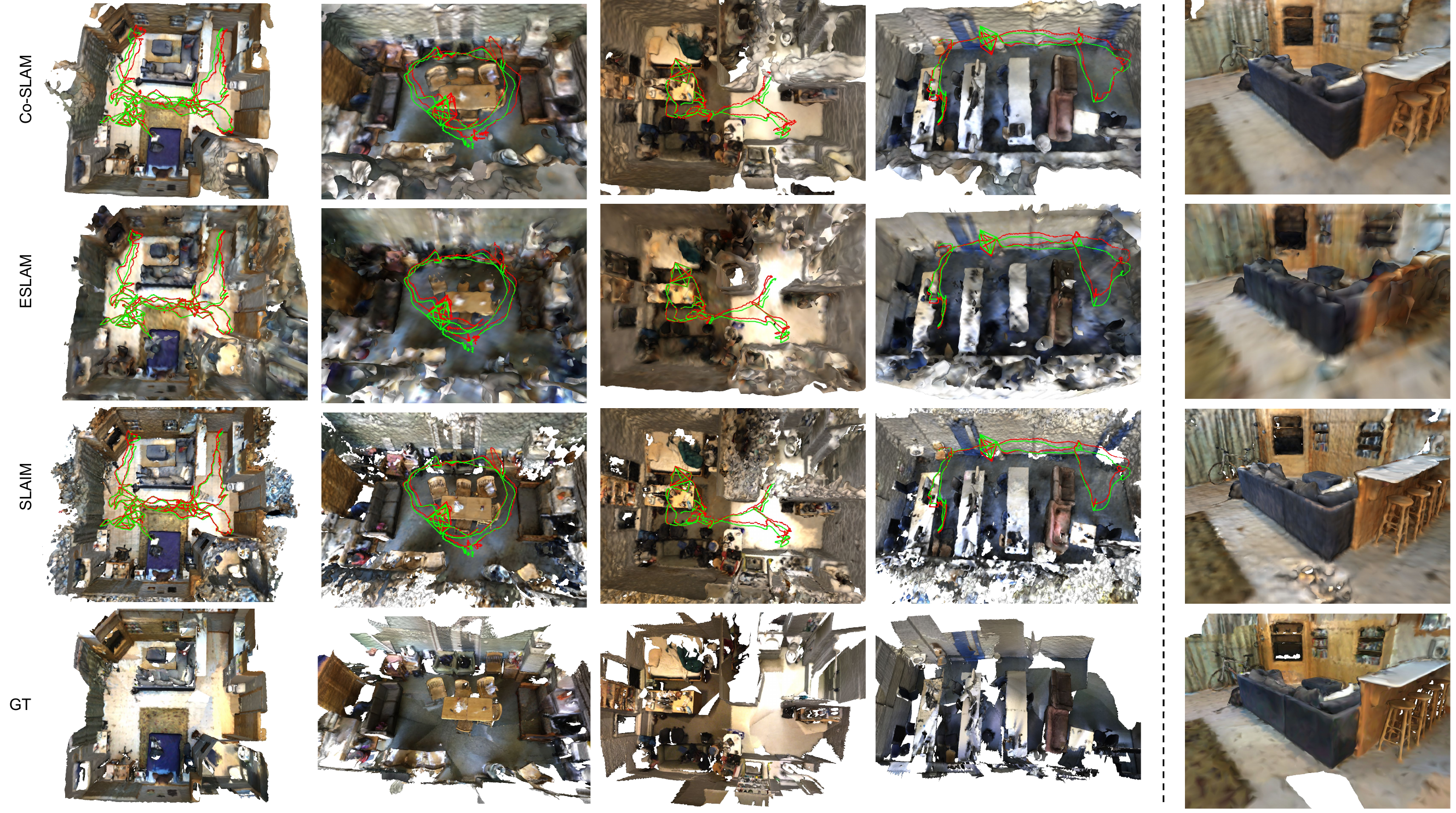}
    \caption{SLAIM qualitative results on the ScanNet dataset \cite{dai2017scannet}. The ground truth camera trajectory is shown in green, and the estimated one in red. In comparison to other baselines \cite{johari2022eslam, wang2023co} our method achieves more accurate tracking results and produces finer 3D reconstruction (right column). 
    }
    \label{fig:results_figure}
\end{figure*}

\textbf{ScanNet dataset.} We evaluate the camera tracking accuracy of SLAIM on 6 real-world sequences from ScanNet \cite{dai2017scannet}. We compute the absolute trajectory error (ATE) between the aligned prediction and ground-truth trajectories. Tab. \ref{tab:results_scannet} shows that quantitatively, our method achieves the best tracking results across the board on all scenes with an increase of close to $15\%$ on average.

\begin{table}[h]
\centering
\renewcommand{\arraystretch}{1.15}
\resizebox{0.95\columnwidth}{!}{
\begin{tabular}{ l c c c c c c c c}
\toprule
scene-ID & & 0000 & 0059 & 0106 & 0169 & 0181 & 0207 & Avg. \\ 
\midrule
iMAP*\cite{sucar2021imap}     & & 55.95 & 32.06 & 17.50 & 70.51 & 32.10 & 11.91 & 36.67 \\
NICE-SLAM \cite{zhu2022nice} & & 8.64  & 12.25 & 8.09 & 10.28 & 12.93 & 5.59 & 9.63 \\
Co-SLAM \cite{wang2023co}    & & 7.13  & 11.14 & 9.36 & 5.90 & 11.81 & 7.14 & 8.75  \\
ESLAM \cite{johari2022eslam} & & 7.32  & 8.55 & 7.51 & 6.51 & 9.21 & 5.71 & 7.42 \\
\midrule
SLAIM ($Ours$) & & \textbf{4.56} & \textbf{6.12} & \textbf{6.9} & \textbf{5.82} & \textbf{8.88} & \textbf{5.69} & \textbf{6.32} \\
\bottomrule
\end{tabular}}\\[2pt]
\caption{Tracking results, ATE RMSE(cm) $\downarrow$, on the ScanNet dataset \cite{dai2017scannet}. Our method gets the best tracking performances across the board.}
\label{tab:results_scannet}
\end{table}

\textbf{TUM dataset.} We also evaluate the tracking performances on the TUM dataset \cite{sturm2012benchmark}. As depicted in Tab. \ref{tab:results_tum}, our method achieves best tracking results on two of the three scenes, and second best on the third one. We notice that across the board our method outperforms the $SLAIM_{MG}$ baseline which highlights the benefits of performing proper coarse-to-fine with a Gaussian Pyramid filter compared to rendering views with different grid feature resolution. The coarse views rendered by $SLAIM_{MG}$ look blurry at first glance but have artifacts caused by the MLP decoders trying to over-compensate the lack of high-resolution features (see visualizations in supplementary material), which weakens the training signal during the early tracking iterations.

\begin{table}[h]
\centering
\resizebox{0.8\columnwidth}{!}{
\begin{tabular}{ l c c c c c}
\toprule
& & fr1/desk & fr2/xyz & fr3/office \\ 
\midrule
iMAP\cite{sucar2021imap}     & & 4.9 & 2.0 & 5.8 \\
NICE-SLAM \cite{zhu2022nice} & & 2.7 & 1.8 & 3.0 \\
Co-SLAM \cite{wang2023co}    & & 2.4 & 1.7 & 2.4  \\
ESLAM \cite{johari2022eslam} & & 2.5 & \textbf{1.1} & 2.4  \\
\midrule
$SLAIM (Ours)$ & & \textbf{2.1} & 1.5 & \textbf{2.3} \\
$SLAIM_{MG}$ & & 2.5 & 1.8 & 2.5 \\
\midrule
ORB-SLAM2 \cite{mur2017orb} & & 1.6 & 0.4 & 1.0 \\
\bottomrule
\end{tabular}}\\[2pt]
\caption{RMSE ATE(cm) $\downarrow$ tracking results on the TUM-RGBD dataset \cite{sturm2012benchmark}. SLAIM gets the best performances on two of the three scenes and second best on the third scene.}
\label{tab:results_tum}
\end{table}

\subsection{Ablation studies.}


\textbf{Effect of the coarse-to-fine strategy on Mapping and Tracking.} We report tracking accuracy (ATE) numbers on the ScanNet dataset \cite{dai2017scannet} in Tab. \ref{tab:results_ablation_c2f} showing a significant performance increase when using the coarse-to-fine strategy on both mapping and tracking. We compare to a baseline where no coarse-to-fine is applied ($SLAIM_{noc2f}$) and observe an $8\%$ increase in performances when using coarse-to-fine (lines 1 and 5). The $SLAIM_{track}$ and $SLAIM_{map}$ are baselines where we run the coarse-to-fine strategy on the tracker or mapper only. We notice that coarse-to-fine tracking has a bigger impact than coarse-to-fine mapping (lines 2-3). 

\begin{table}[h]
\centering
\resizebox{0.95\columnwidth}{!}{
\begin{tabular}{ l c c c c c c c}
\toprule
        & & Map. c2f & Track. c2f & GPL & & AVG ATE(cm) \\ 
\midrule
$SLAIM_{noc2f}$ & &  &  & N/A & & 6.91 \\
$SLAIM_{track}$ & &  & \checkmark & 2 & & 6.56 \\
$SLAIM_{map}$ & & \checkmark &  & 2 & & 6.90 \\
\midrule
$SLAIM$ & & \checkmark & \checkmark & 1 & & 6.44 \\
$SLAIM$ & & \checkmark & \checkmark & 2 & & \textbf{6.32} \\
$SLAIM$ & & \checkmark & \checkmark & 3 & & 6.47 \\
\bottomrule
\end{tabular}}\\[2pt]
\caption{Results on ScanNet \cite{dai2017scannet} showing the impact of the coarse-to-fine (c2f) strategy along with the number of Gaussian Pyramid levels. We observe best performances when doing c2f with a level-2 Gaussian Pyramid compared to no c2f ($noc2f$) and baselines that only do c2f during tracking ($track$) and mapping ($map$).}
\label{tab:results_ablation_c2f}
\end{table}

\textbf{Impact of the number of levels in the Gaussian Pyramid}
We compare the tracking accuracy on the ScanNet dataset in Tab. \ref{tab:results_ablation_c2f} when testing with different GPL. From Tab. \ref{tab:results_ablation_c2f} we observe that the optimal GPL is 2 (lines 4-6). Increasing the GPL to 3 decreases the performances. We explain this phenomenon by the fact that the receptive field of the Gaussian Pyramid increases as the GPL increases. Therefore, with a fixed set of rendered rays per batch, the number of pixels at the GPL on which the loss is computed will reduce as the GPL increases.

\textbf{Geometry is important.}
Tab. \ref{tab:results_ablation_BA} reports ablation tracking accuracy results on the ScanNet \cite{dai2017scannet} dataset. We observe that our approach with the custom KL regularizer (line 3) yields the best results with an increase of close to $5\%$ in average ATE compared to $SLAIM_G$ which uses the Gaussian regularizer implemented in DSNerf \cite{deng2022depth}. We also observe that the results drop drastically when not using any KL regularizer ($SLAIM_{noKL}$ - line 5).

\begin{table}[h]
\centering
\resizebox{0.95\columnwidth}{!}{
\begin{tabular}{ l c c c c c c c}
\toprule
        & & BA & KL reg. & & AVG ATE(cm) \\ 
\midrule
LBA & & L & $Ours$ & & 9.45 \\
GBA & & G & $Ours$ & & 11.32 \\
$SLAIM$ & & G+L & $Ours$ & & \textbf{6.32} \\
$SLAIM_{G}$ & & G+L & Gaussian & & 6.69 \\
$SLAIM_{noKL}$ & & G+L & N/A & & 9.21 \\
\bottomrule
\end{tabular}}\\[2pt]
\caption{Ablation study showing the impact of local (L) and global (G) bundle adjustment (BA) and the choice of the KL regularizer. We get the best performances when doing both global and local bundle-adjustment and using our custom KL regularizer.}
\label{tab:results_ablation_BA}
\end{table}

\textbf{Local and global bundle adjustments.}
From Tab. \ref{tab:results_ablation_BA} we measure the impact of the local and global bundle-adjustment. We observe that we obtain the best results when combining both global and local bundle-adjustment (line 3). We notice a large drop in performances when doing local (LBA) or global (GBA) bundle-adjustment only (lines 1,2).

\begin{table}[h]
\centering
\renewcommand{\arraystretch}{1.15}
\resizebox{0.95\columnwidth}{!}{
\begin{tabular}{ c | l | c c c c}
\toprule
& & Track. (ms) $\downarrow$ & Map. (ms) $\downarrow$ & FPS $\uparrow$ & $\#$Params $\downarrow$  \\ 
\midrule
\multirow{5}{*}{\rotatebox[origin=c]{90}{Replica}} 
 & iMAP & $21.1\times6$ & $46.1\times10$ & $4.5$ & $0.26$ M\\
 & NICE-SLAM & $7.9\times10$ & $91.2\times60$ & $0.98$ & $17.4$ M\\
 & Co-SLAM & $5.9\times 10$ & $10.1\times 10$ & $12.6$& $0.26$ M\\
 & ESLAM & $7.5\times 8$ & $19.9\times 15$ & $6.4$ & $9.29$ M\\
& $SLAIM$ (Ours) & $7.4\times10$ & $9.8\times 20$ & $8.8$ & $0.25$ M\\
\bottomrule
\multirow{5}{*}{\rotatebox[origin=c]{90}{ScanNet}}
 & iMAP & $30.4\times50$ & $44.9\times300$ & $0.37$ & $0.2$ M \\
 & NICE-SLAM & $12.3 \times 50$ & $125.3\times60$ & $0.68$ & $10.3$ M\\
 & Co-SLAM & $7.8 \times 10$ & $20.2\times 10$ & $8.4$ & $0.8$ M\\
& ESLAM & $7.4 \times 30$ & $22.4 \times 30 $ & $2.81$ & $10.5$ M\\
& $SLAIM$ (Ours) & $8.4 \times 15$ & $23.1\times (15+15)$ & $3.8$ & $0.8$ M\\
\end{tabular}}\\[2pt]
\caption{Run time and memory footprint comparison across baselines on Replica and ScanNet datasets. Track.(ms) and Map. (ms) are the tracking and mapping time shown as the time per iteration multiplied by the number of iterations. For SLAIM the number of mapping iterations comprises the local and global bundle-adjustment iterations (we do only global BA on Replica).}
\label{tab:runtime}
\end{table}

\subsection{Run time comparison.}
Tab. \ref{tab:runtime} reports the run times and memory footprint of the different baselines on Replica and ScanNet. Our method has similar FPS compared to ESLAM while using significantly less memory - close to ten times less. In addition, SLAIM presents better tracking and 3D reconstruction results compared to Co-SLAM at the price of a lower FPS. 

\section{Conclusion}
\label{sec:conclusion}

We introduced SLAIM, a dense real-time RGB-D NeRF-SLAM system with state-of-the-art camera tracking and mapping. We show that using coarse-to-fine tracking and photometric bundle-adjustment along with proper depth supervision, SLAIM achieves both accurate camera tracking and high-quality 3D reconstruction while staying memory efficient. We show from extensive experiments, comparisons, and ablations that our Gaussian Pyramid filter and our new KL regularizer lead to SOTA results in camera tracking and 3D reconstruction.

{
    \small
    \bibliographystyle{ieeenat_fullname}
    \bibliography{main}

\begin{thebibliography}{42}
\providecommand{\natexlab}[1]{#1}
\providecommand{\url}[1]{\texttt{#1}}
\expandafter\ifx\csname urlstyle\endcsname\relax
  \providecommand{\doi}[1]{doi: #1}\else
  \providecommand{\doi}{doi: \begingroup \urlstyle{rm}\Url}\fi

\bibitem[Azinovi{\'c} et~al.(2022)Azinovi{\'c}, Martin-Brualla, Goldman, Nie{\ss}ner, and Thies]{azinovic2022neural}
Dejan Azinovi{\'c}, Ricardo Martin-Brualla, Dan~B Goldman, Matthias Nie{\ss}ner, and Justus Thies.
\newblock Neural rgb-d surface reconstruction.
\newblock In \emph{Proceedings of the IEEE/CVF Conference on Computer Vision and Pattern Recognition}, pages 6290--6301, 2022.

\bibitem[Burt and Adelson(1987)]{burt1987laplacian}
Peter~J Burt and Edward~H Adelson.
\newblock The laplacian pyramid as a compact image code.
\newblock In \emph{Readings in computer vision}, pages 671--679. Elsevier, 1987.

\bibitem[Bylow et~al.(2013)Bylow, Sturm, Kerl, Kahl, and Cremers]{bylow2013real}
Erik Bylow, J{\"u}rgen Sturm, Christian Kerl, Fredrik Kahl, and Daniel Cremers.
\newblock Real-time camera tracking and 3d reconstruction using signed distance functions.
\newblock In \emph{Robotics: Science and Systems}, page~2, 2013.

\bibitem[Cadena et~al.(2016)Cadena, Carlone, Carrillo, Latif, Scaramuzza, Neira, Reid, and Leonard]{cadena2016past}
Cesar Cadena, Luca Carlone, Henry Carrillo, Yasir Latif, Davide Scaramuzza, Jos{\'e} Neira, Ian Reid, and John~J Leonard.
\newblock Past, present, and future of simultaneous localization and mapping: Toward the robust-perception age.
\newblock \emph{IEEE Transactions on robotics}, 32\penalty0 (6):\penalty0 1309--1332, 2016.

\bibitem[Campos et~al.(2021)Campos, Elvira, Rodr{\'\i}guez, Montiel, and Tard{\'o}s]{campos2021orb}
Carlos Campos, Richard Elvira, Juan J~G{\'o}mez Rodr{\'\i}guez, Jos{\'e}~MM Montiel, and Juan~D Tard{\'o}s.
\newblock Orb-slam3: An accurate open-source library for visual, visual--inertial, and multimap slam.
\newblock \emph{IEEE Transactions on Robotics}, 37\penalty0 (6):\penalty0 1874--1890, 2021.

\bibitem[Chabra et~al.(2020)Chabra, Lenssen, Ilg, Schmidt, Straub, Lovegrove, and Newcombe]{chabra2020deep}
Rohan Chabra, Jan~E Lenssen, Eddy Ilg, Tanner Schmidt, Julian Straub, Steven Lovegrove, and Richard Newcombe.
\newblock Deep local shapes: Learning local sdf priors for detailed 3d reconstruction.
\newblock In \emph{European Conference on Computer Vision}, pages 608--625. Springer, 2020.

\bibitem[Dai et~al.(2017{\natexlab{a}})Dai, Chang, Savva, Halber, Funkhouser, and Nie{\ss}ner]{dai2017scannet}
Angela Dai, Angel~X Chang, Manolis Savva, Maciej Halber, Thomas Funkhouser, and Matthias Nie{\ss}ner.
\newblock Scannet: Richly-annotated 3d reconstructions of indoor scenes.
\newblock In \emph{Proceedings of the IEEE conference on computer vision and pattern recognition}, pages 5828--5839, 2017{\natexlab{a}}.

\bibitem[Dai et~al.(2017{\natexlab{b}})Dai, Nie{\ss}ner, Zollh{\"o}fer, Izadi, and Theobalt]{dai2017bundlefusion}
Angela Dai, Matthias Nie{\ss}ner, Michael Zollh{\"o}fer, Shahram Izadi, and Christian Theobalt.
\newblock Bundlefusion: Real-time globally consistent 3d reconstruction using on-the-fly surface reintegration.
\newblock \emph{ACM Transactions on Graphics (ToG)}, 36\penalty0 (4):\penalty0 1, 2017{\natexlab{b}}.

\bibitem[Dai et~al.(2020)Dai, Diller, and Nie{\ss}ner]{dai2020sg}
Angela Dai, Christian Diller, and Matthias Nie{\ss}ner.
\newblock Sg-nn: Sparse generative neural networks for self-supervised scene completion of rgb-d scans.
\newblock In \emph{Proceedings of the IEEE/CVF Conference on Computer Vision and Pattern Recognition}, pages 849--858, 2020.

\bibitem[Deng et~al.(2022)Deng, Liu, Zhu, and Ramanan]{deng2022depth}
Kangle Deng, Andrew Liu, Jun-Yan Zhu, and Deva Ramanan.
\newblock Depth-supervised nerf: Fewer views and faster training for free.
\newblock In \emph{Proceedings of the IEEE/CVF Conference on Computer Vision and Pattern Recognition}, pages 12882--12891, 2022.

\bibitem[Fridovich-Keil et~al.(2022)Fridovich-Keil, Yu, Tancik, Chen, Recht, and Kanazawa]{fridovich2022plenoxels}
Sara Fridovich-Keil, Alex Yu, Matthew Tancik, Qinhong Chen, Benjamin Recht, and Angjoo Kanazawa.
\newblock Plenoxels: Radiance fields without neural networks.
\newblock In \emph{Proceedings of the IEEE/CVF Conference on Computer Vision and Pattern Recognition}, pages 5501--5510, 2022.

\bibitem[Johari et~al.(2022)Johari, Carta, and Fleuret]{johari2022eslam}
Mohammad~Mahdi Johari, Camilla Carta, and Fran{\c{c}}ois Fleuret.
\newblock Eslam: Efficient dense slam system based on hybrid representation of signed distance fields.
\newblock \emph{arXiv preprint arXiv:2211.11704}, 2022.

\bibitem[K{\"a}hler et~al.(2016)K{\"a}hler, Prisacariu, and Murray]{kahler2016real}
Olaf K{\"a}hler, Victor~A Prisacariu, and David~W Murray.
\newblock Real-time large-scale dense 3d reconstruction with loop closure.
\newblock In \emph{European Conference on Computer Vision}, pages 500--516. Springer, 2016.

\bibitem[Kingma and Ba(2014)]{kingma2014adam}
Diederik~P Kingma and Jimmy Ba.
\newblock Adam: A method for stochastic optimization.
\newblock \emph{arXiv preprint arXiv:1412.6980}, 2014.

\bibitem[Klein and Murray(2009)]{klein2009parallel}
Georg Klein and David Murray.
\newblock Parallel tracking and mapping on a camera phone.
\newblock In \emph{2009 8th IEEE International Symposium on Mixed and Augmented Reality}, pages 83--86. IEEE, 2009.

\bibitem[Kong et~al.(2023)Kong, Liu, Taher, and Davison]{kong2023vmap}
Xin Kong, Shikun Liu, Marwan Taher, and Andrew~J Davison.
\newblock vmap: Vectorised object mapping for neural field slam.
\newblock \emph{arXiv preprint arXiv:2302.01838}, 2023.

\bibitem[Lin et~al.(2021)Lin, Ma, Torralba, and Lucey]{lin2021barf}
Chen-Hsuan Lin, Wei-Chiu Ma, Antonio Torralba, and Simon Lucey.
\newblock Barf: Bundle-adjusting neural radiance fields.
\newblock In \emph{Proceedings of the IEEE/CVF International Conference on Computer Vision}, pages 5741--5751, 2021.

\bibitem[Lucas et~al.(1981)Lucas, Kanade, et~al.]{lucas1981iterative}
Bruce~D Lucas, Takeo Kanade, et~al.
\newblock \emph{An iterative image registration technique with an application to stereo vision}.
\newblock Vancouver, 1981.

\bibitem[Mildenhall et~al.(2021)Mildenhall, Srinivasan, Tancik, Barron, Ramamoorthi, and Ng]{mildenhall2021nerf}
Ben Mildenhall, Pratul~P Srinivasan, Matthew Tancik, Jonathan~T Barron, Ravi Ramamoorthi, and Ren Ng.
\newblock Nerf: Representing scenes as neural radiance fields for view synthesis.
\newblock \emph{Communications of the ACM}, 65\penalty0 (1):\penalty0 99--106, 2021.

\bibitem[M{\"u}ller et~al.(2022)M{\"u}ller, Evans, Schied, and Keller]{muller2022instant}
Thomas M{\"u}ller, Alex Evans, Christoph Schied, and Alexander Keller.
\newblock Instant neural graphics primitives with a multiresolution hash encoding.
\newblock \emph{arXiv preprint arXiv:2201.05989}, 2022.

\bibitem[Mur-Artal and Tard{\'o}s(2017)]{mur2017orb}
Raul Mur-Artal and Juan~D Tard{\'o}s.
\newblock Orb-slam2: An open-source slam system for monocular, stereo, and rgb-d cameras.
\newblock \emph{IEEE transactions on robotics}, 33\penalty0 (5):\penalty0 1255--1262, 2017.

\bibitem[Mur-Artal et~al.(2015)Mur-Artal, Montiel, and Tardos]{mur2015orb}
Raul Mur-Artal, Jose Maria~Martinez Montiel, and Juan~D Tardos.
\newblock Orb-slam: a versatile and accurate monocular slam system.
\newblock \emph{IEEE transactions on robotics}, 31\penalty0 (5):\penalty0 1147--1163, 2015.

\bibitem[Newcombe et~al.(2011{\natexlab{a}})Newcombe, Izadi, Hilliges, Molyneaux, Kim, Davison, Kohi, Shotton, Hodges, and Fitzgibbon]{newcombe2011kinectfusion}
Richard~A Newcombe, Shahram Izadi, Otmar Hilliges, David Molyneaux, David Kim, Andrew~J Davison, Pushmeet Kohi, Jamie Shotton, Steve Hodges, and Andrew Fitzgibbon.
\newblock Kinectfusion: Real-time dense surface mapping and tracking.
\newblock In \emph{2011 10th IEEE international symposium on mixed and augmented reality}, pages 127--136. Ieee, 2011{\natexlab{a}}.

\bibitem[Newcombe et~al.(2011{\natexlab{b}})Newcombe, Lovegrove, and Davison]{newcombe2011dtam}
Richard~A Newcombe, Steven~J Lovegrove, and Andrew~J Davison.
\newblock Dtam: Dense tracking and mapping in real-time.
\newblock In \emph{2011 international conference on computer vision}, pages 2320--2327. IEEE, 2011{\natexlab{b}}.

\bibitem[Or-El et~al.(2022)Or-El, Luo, Shan, Shechtman, Park, and Kemelmacher-Shlizerman]{or2022stylesdf}
Roy Or-El, Xuan Luo, Mengyi Shan, Eli Shechtman, Jeong~Joon Park, and Ira Kemelmacher-Shlizerman.
\newblock Stylesdf: High-resolution 3d-consistent image and geometry generation.
\newblock In \emph{Proceedings of the IEEE/CVF Conference on Computer Vision and Pattern Recognition}, pages 13503--13513, 2022.

\bibitem[Peng et~al.(2020)Peng, Niemeyer, Mescheder, Pollefeys, and Geiger]{peng2020convolutional}
Songyou Peng, Michael Niemeyer, Lars Mescheder, Marc Pollefeys, and Andreas Geiger.
\newblock Convolutional occupancy networks.
\newblock In \emph{European Conference on Computer Vision}, pages 523--540. Springer, 2020.

\bibitem[Rosinol et~al.(2020)Rosinol, Abate, Chang, and Carlone]{rosinol2020kimera}
Antoni Rosinol, Marcus Abate, Yun Chang, and Luca Carlone.
\newblock Kimera: an open-source library for real-time metric-semantic localization and mapping.
\newblock In \emph{2020 IEEE International Conference on Robotics and Automation (ICRA)}, pages 1689--1696. IEEE, 2020.

\bibitem[Rosinol et~al.(2022)Rosinol, Leonard, and Carlone]{rosinol2022nerf}
Antoni Rosinol, John~J Leonard, and Luca Carlone.
\newblock Nerf-slam: Real-time dense monocular slam with neural radiance fields.
\newblock \emph{arXiv preprint arXiv:2210.13641}, 2022.

\bibitem[Schops et~al.(2019)Schops, Sattler, and Pollefeys]{schops2019bad}
Thomas Schops, Torsten Sattler, and Marc Pollefeys.
\newblock Bad slam: Bundle adjusted direct rgb-d slam.
\newblock In \emph{Proceedings of the IEEE/CVF Conference on Computer Vision and Pattern Recognition}, pages 134--144, 2019.

\bibitem[Straub et~al.(2019)Straub, Whelan, Ma, Chen, Wijmans, Green, Engel, Mur-Artal, Ren, Verma, et~al.]{straub2019replica}
Julian Straub, Thomas Whelan, Lingni Ma, Yufan Chen, Erik Wijmans, Simon Green, Jakob~J Engel, Raul Mur-Artal, Carl Ren, Shobhit Verma, et~al.
\newblock The replica dataset: A digital replica of indoor spaces.
\newblock \emph{arXiv preprint arXiv:1906.05797}, 2019.

\bibitem[Sturm et~al.(2012)Sturm, Engelhard, Endres, Burgard, and Cremers]{sturm2012benchmark}
J{\"u}rgen Sturm, Nikolas Engelhard, Felix Endres, Wolfram Burgard, and Daniel Cremers.
\newblock A benchmark for the evaluation of rgb-d slam systems.
\newblock In \emph{2012 IEEE/RSJ international conference on intelligent robots and systems}, pages 573--580. IEEE, 2012.

\bibitem[Sucar et~al.(2021)Sucar, Liu, Ortiz, and Davison]{sucar2021imap}
Edgar Sucar, Shikun Liu, Joseph Ortiz, and Andrew~J Davison.
\newblock imap: Implicit mapping and positioning in real-time.
\newblock In \emph{Proceedings of the IEEE/CVF International Conference on Computer Vision}, pages 6229--6238, 2021.

\bibitem[Takikawa et~al.(2021)Takikawa, Litalien, Yin, Kreis, Loop, Nowrouzezahrai, Jacobson, McGuire, and Fidler]{takikawa2021neural}
Towaki Takikawa, Joey Litalien, Kangxue Yin, Karsten Kreis, Charles Loop, Derek Nowrouzezahrai, Alec Jacobson, Morgan McGuire, and Sanja Fidler.
\newblock Neural geometric level of detail: Real-time rendering with implicit 3d shapes.
\newblock In \emph{Proceedings of the IEEE/CVF Conference on Computer Vision and Pattern Recognition}, pages 11358--11367, 2021.

\bibitem[Tang et~al.(2020)Tang, Singh, Chou, Hane, Dou, Fanello, Taylor, Davidson, Guleryuz, Zhang, et~al.]{tang2020deep}
Danhang Tang, Saurabh Singh, Philip~A Chou, Christian Hane, Mingsong Dou, Sean Fanello, Jonathan Taylor, Philip Davidson, Onur~G Guleryuz, Yinda Zhang, et~al.
\newblock Deep implicit volume compression.
\newblock In \emph{Proceedings of the IEEE/CVF conference on computer vision and pattern recognition}, pages 1293--1303, 2020.

\bibitem[Teed and Deng(2021)]{teed2021droid}
Zachary Teed and Jia Deng.
\newblock Droid-slam: Deep visual slam for monocular, stereo, and rgb-d cameras.
\newblock \emph{Advances in Neural Information Processing Systems}, 34:\penalty0 16558--16569, 2021.

\bibitem[Wang et~al.(2023)Wang, Wang, and Agapito]{wang2023co}
Hengyi Wang, Jingwen Wang, and Lourdes Agapito.
\newblock Co-slam: Joint coordinate and sparse parametric encodings for neural real-time slam.
\newblock \emph{arXiv preprint arXiv:2304.14377}, 2023.

\bibitem[Wang et~al.(2021{\natexlab{a}})Wang, Liu, Liu, Theobalt, Komura, and Wang]{wang2021neus}
Peng Wang, Lingjie Liu, Yuan Liu, Christian Theobalt, Taku Komura, and Wenping Wang.
\newblock Neus: Learning neural implicit surfaces by volume rendering for multi-view reconstruction.
\newblock \emph{arXiv preprint arXiv:2106.10689}, 2021{\natexlab{a}}.

\bibitem[Wang et~al.(2021{\natexlab{b}})Wang, Wu, Xie, Chen, and Prisacariu]{wang2021nerf}
Zirui Wang, Shangzhe Wu, Weidi Xie, Min Chen, and Victor~Adrian Prisacariu.
\newblock Nerf--: Neural radiance fields without known camera parameters.
\newblock \emph{arXiv preprint arXiv:2102.07064}, 2021{\natexlab{b}}.

\bibitem[Whelan et~al.(2015)Whelan, Leutenegger, Salas-Moreno, Glocker, and Davison]{whelan2015elasticfusion}
Thomas Whelan, Stefan Leutenegger, Renato Salas-Moreno, Ben Glocker, and Andrew Davison.
\newblock Elasticfusion: Dense slam without a pose graph.
\newblock Robotics: Science and Systems, 2015.

\bibitem[Yariv et~al.(2021)Yariv, Gu, Kasten, and Lipman]{yariv2021volume}
Lior Yariv, Jiatao Gu, Yoni Kasten, and Yaron Lipman.
\newblock Volume rendering of neural implicit surfaces.
\newblock \emph{Advances in Neural Information Processing Systems}, 34:\penalty0 4805--4815, 2021.

\bibitem[Yariv et~al.(2023)Yariv, Hedman, Reiser, Verbin, Srinivasan, Szeliski, Barron, and Mildenhall]{yariv2023bakedsdf}
Lior Yariv, Peter Hedman, Christian Reiser, Dor Verbin, Pratul~P Srinivasan, Richard Szeliski, Jonathan~T Barron, and Ben Mildenhall.
\newblock Bakedsdf: Meshing neural sdfs for real-time view synthesis.
\newblock \emph{arXiv preprint arXiv:2302.14859}, 2023.

\bibitem[Zhu et~al.(2022)Zhu, Peng, Larsson, Xu, Bao, Cui, Oswald, and Pollefeys]{zhu2022nice}
Zihan Zhu, Songyou Peng, Viktor Larsson, Weiwei Xu, Hujun Bao, Zhaopeng Cui, Martin~R Oswald, and Marc Pollefeys.
\newblock Nice-slam: Neural implicit scalable encoding for slam.
\newblock In \emph{Proceedings of the IEEE/CVF Conference on Computer Vision and Pattern Recognition}, pages 12786--12796, 2022.

\end{thebibliography}
}

\newpage
\section{Additional implementation details}
\label{sec:impl_details}

\subsection{Derivation of the KL regularizer}
In order to model the ray termination distribution we start by representing the density $\tilde{\sigma}$ as a narrow bell shaped function using the $sech^2$ function. We use the $sech^2$ over a Gaussian for its simplicity when computing integrals. We define the estimated density as follow:
\begin{equation}
    \tilde{\sigma}(r) = S\times sech^2 \left( \frac{(r-d')}{\sigma_d} \right)
\end{equation}

With $d'$ and $\sigma_d$ the mean and variance and $S$ a scale factor. The final ray termination distribution is then computed using the following equation: $\tilde{w}(r) = T(r)\tilde{\sigma}(r)$, with $T(r) = exp(-\int_{0}^{r} \tilde{\sigma}(s) ds)$. We intent to have an analytical formulation of $\tilde{w}(r)$ so it is easier and faster to compute the resulting KL regularizer. In order to do this we need to compute the integral in $T(r)$, $\int_{0}^{r} \tilde{\sigma}(s) ds$. We start with the following,
\begin{equation} \label{eq:sech_int}
    \begin{split}
      \int_{0}^{r} \tilde{\sigma}(s) ds & = \int_{0}^{r} S\times sech^2 \left( \frac{(s-d')}{\sigma_d} \right) ds \\
           & =  S \sigma_d \left[ tanh\left( \frac{(s-d')}{\sigma_d} \right) \right]^{r}_0 \\
           & =  S \sigma_d \left[ tanh\left( \frac{(r-d')}{\sigma_d} \right) - tanh\left( \frac{-d'}{\sigma_d} \right) \right] 
\end{split}
\end{equation}

We can plug the result of Eq \ref{eq:sech_int} into our custom ray termination distribution and express $\tilde{w}$ analytically as follow

\begin{equation}\label{eq:w}
    \begin{split}
        \tilde{w}(r) & = \tilde{\sigma}(r)T(r)\\
                    & = \tilde{\sigma}(r) e^{-S \sigma_d \left[ tanh\left( \frac{(r-d')}{\sigma_d} \right) - tanh\left( \frac{-d'}{\sigma_d} \right) \right]} \\
                    & = S. sech^2 \left( \frac{(r-d')}{\sigma_d} \right) e^{-S \sigma_d \left[ tanh\left( \frac{(r-d')}{\sigma_d} \right) - tanh\left( \frac{-d'}{\sigma_d} \right) \right]}
    \end{split}
\end{equation}

We can now use that estimation of a ray termination distribution in a KL regularizer. 
\begin{equation}
\begin{split}
      \mathcal{L}_{KL} & =  -\frac{1}{N}\sum^{N}_{n=1} \sum_k log(w(r_k))\tilde{w}(r_k) \Delta_r
\end{split}
\end{equation}

With $N$ the number of rays per batch, $\Delta_r$ the ray marching step size and $\tilde{w}$ the predicted ray termination distribution. Note here that although a we could have represented the $\tilde{\sigma}$ function with a Gaussian, we wouldn't have been able to compute an analytical formulation of the integral on Eq. \ref{eq:sech_int}, and estimating such integral would have been computationally costly.

\subsection{Expectation of the custom ray termination distribution}
It is important that the expectation of the ray termination distribution be centered on the depth measurement $d$ in order to ensure a good quality of reconstruction. The custom ray termination distriubtion involves a mean $d'$ and variance $\sigma_d$ parameters. Naively setting $d'=d$ would lead to a bias estimator. We can compute the expected value of the distribution $\tilde{w}$ in order to set the parameter $d'$. 

\begin{equation}
    \begin{split}
        E_r(\tilde{w}) & = \int_{-\infty}^{\infty} r.\tilde{w}(r) dr
    \end{split}
\end{equation}

When reusing the results in Eq. \ref{eq:w} and by setting $y=\frac{(r-d')}{\sigma_d}$ and $C=tanh\left(\frac{-d'}{\sigma_d} \right)$ we have,

\begin{equation}
    \begin{split}
        E_r(\tilde{w}) & = \int_{-\infty}^{\infty} \sigma_d (y\sigma_d + d').S. sech^2 (y) e^{-S \sigma_d \left[ tanh(y) - C \right]}dy \\
        & =  S \sigma_d^2 \int_{-\infty}^{\infty} y . sech^2 (y) e^{-S \sigma_d \left[ tanh(y) - C \right]}dy \\
        & + S \sigma_d d' \int_{-\infty}^{\infty} sech^2 (y) e^{-S \sigma_d \left[ tanh(y) - C \right]}dy\\
        & = S \sigma_d^2 \times \mathcal{A} + S \sigma_d d' \times \mathcal{B}
    \end{split}
\end{equation}

With $\mathcal{A}$ and $\mathcal{B}$ the two integrals. In addition, we can notice that $d' \gg \sigma_d$ therefore we can simplify $C = tanh\left(\frac{-d'}{\sigma_d} \right) \approx -1$. We can now rewrite the two integrals $\mathcal{A}$ and $\mathcal{B}$ as two functions that do not depend on $d'$.
\begin{equation}
    \begin{split}
        \mathcal{A} & = \int_{-\infty}^{\infty} y . sech^2 (y) e^{-S \sigma_d \left[ tanh(y) + 1 \right]}dy \\
        \mathcal{B} & = \int_{-\infty}^{\infty} sech^2 (y) e^{-S \sigma_d \left[ tanh(y) + 1\right]}dy
    \end{split}
\end{equation}
In practice computing these integrals is difficult. However, we only need to evaluate them once. Therefore, we estimate these integrals using a sampling based approach at the beginning of each run. We can then infer the parameter $d'$ by setting the expectation value to the depth measurement $d$, i.e. $E_r(\tilde{w})=d$. We then find,
\begin{equation}
    d' = \frac{d - S \sigma_d^2 \times \mathcal{A} }{S \sigma_d \times \mathcal{B}}
\end{equation}


\subsection{Hyperparameters}
We detail here the hyperparameters and network structure used in SLAIM. We build our code upon the existing Instant-NGP framework \cite{muller2022instant}. Our NeRF model uses hash-grid features with 16 levels and 2 features per levels. The base resolution of the feature grid is set to $R_{min}=16$, and we set the max resolution $R_{max}$ to the next power of 2 value such that it corresponds to a resolution of 1 cm for Replica \cite{straub2019replica}, 2cm for TUM \cite{sturm2012benchmark} and 4cm for ScanNet \cite{dai2017scannet}. We do not use the ray direction as input to our network. Both MLP decoders have one hidden layer with 32 neurons. We use the Adam optimizer \cite{kingma2014adam} with a learning rate of $1.e^{-2}$ without any scheduler. 
In TUM and ScanNet experiments we use a subset of $M=5$ keyframes for local bundle adjustment. We do not perform local bundle adjustment in Replica.


\textbf{TUM settings.} We use an $S=10000$ and $\sigma_d=0.02$ in the custom ray termination distribution. The camera pose parameters' learning rate is set to $1.e^{-3}$ during tracking and $5.e^{-4}$ during mapping. We perform $15$ tracking iterations with a GPL of 2: 5 iterations at each levels 2, 1 and 0. During the mapping phase we perform $15$ iterations of local bundle-adjustment first and $15$ additional global bundle adjustment iterations using a GPL of 2 (ie. 5 iterations at each levels 2 to 0). We sample $N_t=N_m=2048$ rays for both mapping and tracking, we use $\lambda_d=1$ and $\lambda_{KL}=10.$.

\textbf{ScanNet settings.} We use an $S=5000$ and $\sigma_d=0.04$ in the custom ray termination distribution. The camera pose parameters' learning rate is set to $1.e^{-3}$ during tracking and $5.e^{-4}$ during mapping. We perform $15$ tracking iterations with a GPL of 2: 5 iterations at each levels 2, 1 and 0. During the mapping phase we perform $15$ iterations of local bundle-adjustment first and $15$ additional global bundle adjustment iterations using a GPL of 2 (ie. 5 iterations at each levels 2 to 0). We sample $N_t=2048$ and $N_m=4096$ rays for tracking and mapping respectively, and we use $\lambda_d=1$ and $\lambda_{KL}=1.$.

\textbf{Replica settings.} We use an $S=10000$ and $\sigma_d=0.01$ in the custom ray termination distribution. The camera pose parameters' learning rate is set to $1.e^{-3}$ during tracking and $5.e^{-4}$ during mapping. We perform $10$ tracking iterations with a GPL of 1: 5 iterations at each levels 1 and 0. During the mapping phase we perform $20$ iterations of global bundle-adjustment only using a GPL of 1 (ie. 10 iterations at each levels 1 and 0). We sample $N_t=N_m=4096$ rays for tracking and mapping, and we use $\lambda_d=1$ and $\lambda_{KL}=1.$.

\subsection{Camera tracking evaluation protocol}
To evaluate the camera tracking performances we first perform a global alignment between the generated camera poses and the ground-truth ones to alleviate any downgrading effects due to global shifts. This is a common practice in evaluation of SLAM systems \cite{mur2017orb, zhu2022nice, sucar2021imap, wang2023co}. We further compute the absolute translation error (ATE) as the RMSE between the ground-truth poses and the aligned generated ones.

\subsection{3D reconstruction elvaluation protocole}
In the realm of neural implicit reconstruction and SLAM, it is necessary to perform an additional mesh culling step to limit the extrapolation capabilities of NeRf when rendering a mesh outside of the camera view frustrum. We adopt the same strategy as in Co-SLAM \cite{wang2023co}. We compute the global camera viewing frustrum given the set of camera poses plus additional poses generated to handle occlusions. We then remove any vertices outside of that viewing frustrum. This is a simple yet effective technique to limit reconstruction artifacts outside of the targeted region.

\section{Additional details on the $Ours_{MG}$ experiments.}

\begin{figure*}[!htb]
    \centering
    \includegraphics[width=0.85\textwidth]{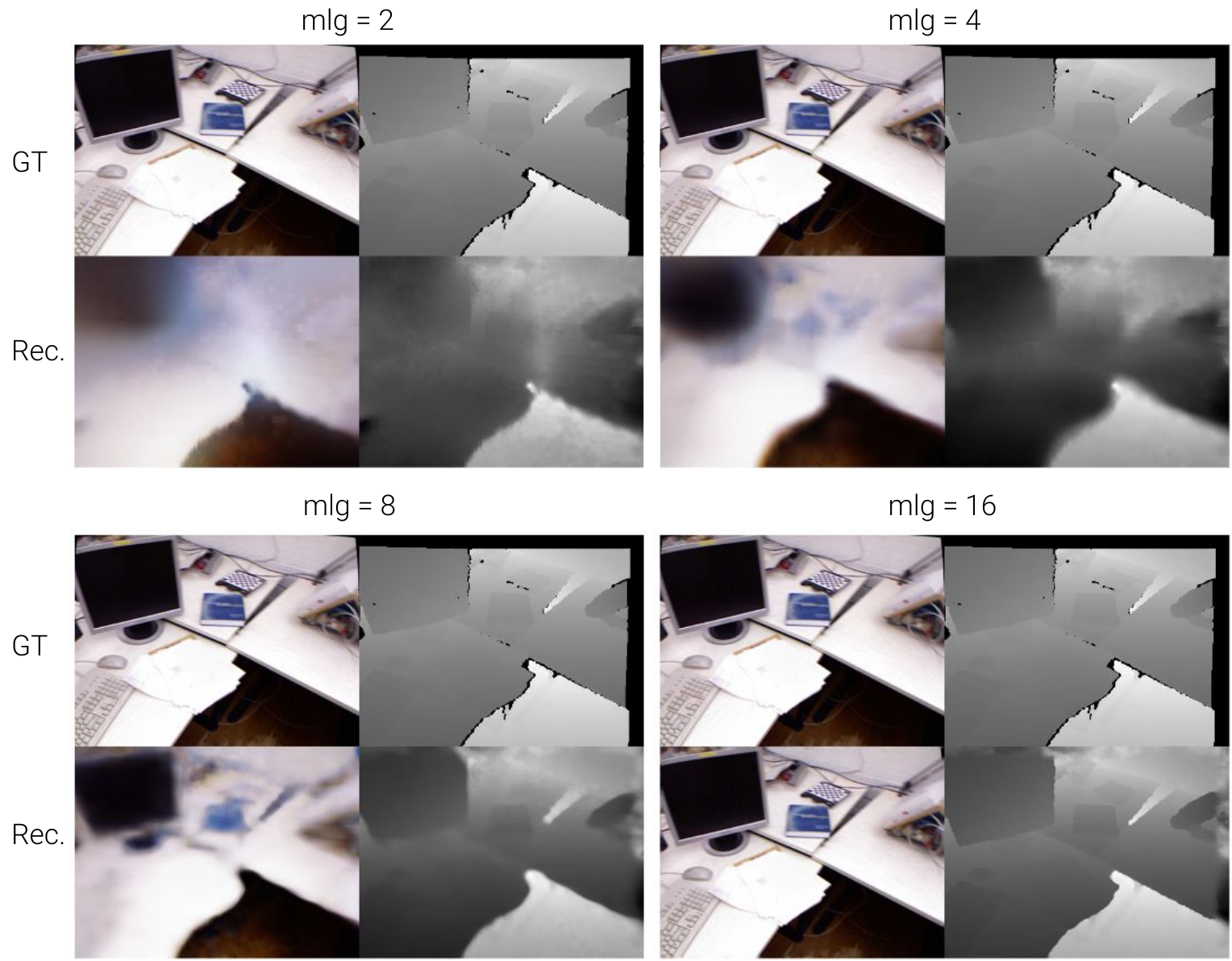}
    \caption{Visualization of 2D NeRF reconstructions of the $SLAIM_{MG}$ model with different $mgl$ levels. Each sub-figures has ground-truth views on the top and reconstructions at the bottom. We show different reconstructions with $mgl$ levels 2, 4, 8, and 16. We observe that the low resolution reconstruction contain artifacts such as the blue area under the monitor in the $mgl=2$ figure. In addition the table edge appears more blurry in $mgl=4$ compared to $mgl=2$ which is counter intuitive. 
    }
    \label{fig:mgl_examples}
\end{figure*}

In our experiments, we compared our SLAIM model to another baseline, that we refer to as $SLAIM_{MG}$ ($MG$ stands for max-grid). This baseline also does coarse-to-fine tracking by rendering blurry images in the early tracking iterations. To construct blurry images $SLAIM_{MG}$ renders images using different grid-resolution features. We use a max-grid-level parameter $mgl \leq L$ that we use to set a max resolution the network is allowed to use to construct the position embedding $y$. Grid features for higher levels are set to $0$, and the final positional encoding can be written as $y = \left[h^{1}_{\beta}(x), ..., h^{mgl}_{\beta}(x), 0 \right]$. This implementation is interesting because it requires no extra training steps or additional FLOPs operations during tracking or mapping.
However, from our results on the TUM dataset \cite{sturm2012benchmark}, we observe that this baseline performs worse than previous baselines. We hypothesis that the reason is because the generated blurry images contain reconstruction artifacts. We show some of these reconstructions defaults in Fig. \ref{fig:mgl_examples}. For instance in the lowest resolution we observe a blue area near the top left part of the image which is inconsistent with the original image. We believe that this phenomena is due to the MLP decoder trying to overcompensate the lack of high-frequency features.

\end{document}